\documentclass[10pt,twocolumn,letterpaper]{article}

\usepackage{iccv}
\usepackage{times}
\usepackage{epsfig}
\usepackage{graphicx}
\usepackage{amsmath}
\usepackage{amssymb}
\usepackage{cite}
\usepackage{soul}
\usepackage{multirow}
\usepackage{color, colortbl}
\usepackage{comment}

\newcommand\blfootnote[1]{%
  \begingroup
  \renewcommand\thefootnote{}\footnote{#1}%
  \addtocounter{footnote}{-1}%
  \endgroup
}

\usepackage[accsupp]{axessibility}

\usepackage[pagebackref=true,breaklinks=true,letterpaper=true,colorlinks,bookmarks=false]{hyperref}

\iccvfinalcopy % *** Uncomment this line for the final submission

\def\iccvPaperID{7095} % *** Enter the ICCV Paper ID here

\ificcvfinal\fi

\begin{document}

%%%%%%%%% TITLE
\title{Event-based Temporally Dense\\Optical Flow Estimation with Sequential Learning}

\author{Wachirawit Ponghiran~~~~~~Chamika Mihiranga Liyanagedera~~~~~~Kaushik Roy\\
Purdue University\\
West Lafayette, IN 47907, USA\\
% \{wponghir,cliyanag,kaushik}\}@purdue.edu
{\tt\small \{wponghir,cliyanag,kaushik\}@purdue.edu}
% \thanks{This work was supported in part by, Center for Brain-inspired Computing (C-BRIC), a DARPA sponsored JUMP center, Semiconductor Research Corporation (SRC), National Science Foundation, the DoD Vannevar Bush Fellowship, and IARPA MicroE4AI.} %
}

\maketitle
\ificcvfinal\thispagestyle{empty}\fi

\blfootnote{This work was supported in part by, Center for Brain-inspired Computing (C-BRIC), a DARPA sponsored JUMP center, Semiconductor Research Corporation (SRC), National Science Foundation, the DoD Vannevar Bush Fellowship, and IARPA MicroE4AI.}
\blfootnote{Code is available at \url{https://github.com/wponghiran/temporally_dense_flow}}

%%%%%%%%% ABSTRACT
\begin{abstract}
Event cameras provide an advantage over traditional frame-based cameras when capturing fast-moving objects without a motion blur. They achieve this by recording changes in light intensity (known as events), thus allowing them to operate at a much higher frequency and making them suitable for capturing motions in a highly dynamic scene. 
Many recent studies have proposed methods to train neural networks (NNs) for predicting optical flow from events. However, they often rely on a spatio-temporal representation constructed from events over a fixed interval, such as 10~Hz used in training on the DSEC dataset.
This limitation restricts the flow prediction to the same interval (10~Hz) whereas the fast speed of event cameras, which can operate up to 3~kHz, has not been effectively utilized.
In this work, we show that a temporally dense flow estimation at 100~Hz can be achieved by treating the flow estimation as a sequential problem using two different variants of recurrent networks -- Long-short term memory (LSTM) and spiking neural network (SNN).
First, We utilize the NN model constructed similar to the popular EV-FlowNet but with LSTM layers to demonstrate the efficiency of our training method. The model not only produces 10$\times$ more frequent optical flow than the existing ones, but the estimated flows also have 13\% lower errors than predictions from the baseline EV-FlowNet. 
Second, we construct an EV-FlowNet SNN but with leaky integrate and fire neurons to efficiently capture the temporal dynamics.
%to show that our proposed method can be extended to train a spiking model as well. 
We found that simple inherent recurrent dynamics of SNN lead to significant parameter reduction compared to the LSTM model. In addition, because of its event-driven computation, the spiking model is estimated to consume only 1.5\% energy of the LSTM model, highlighting the efficiency of SNN in processing events and the potential for achieving temporally dense flow. 
\end{abstract}

%%%%%%%%% BODY TEXT
\section{Introduction}
% Introduction to optical flow prediction and its importance -> Technique to estimate optical with neural networks from frame-based camera outputs -> Shortcoming of frame-based techniques 
Optical flow estimation is a core problem in computer vision that evaluates the motion of each pixel between any two consecutive images captured by a frame-based camera.
Optical flow information enables an observer to visualize a motion field which is useful for numerous applications such as object trajectory prediction~\cite{quan2021holistic}, robotic control~\cite{serres2017optic}, and autonomous driving~\cite{janai2020computer}.
The problem has been traditionally addressed using various classical computer vision techniques like correlation-based~\cite{singh1991optic}, block-matching~\cite{beauchemin1995computation} and energy minimization-based~\cite{horn1981determining} techniques, but their computational costs have shown to be prohibitively expensive for real-time applications. 
Neural network (NN)~based techniques for optical flow prediction  \cite{dosovitskiy2015flownet,ranjan2017optical,sun2018pwc} have been proposed and remain a popular low-cost computing method. 
Generally, NN models receive two consecutive images taken by a frame-based camera as input and predict the optical flow that best warps pixels from one image to another.
However, due to the limited dynamic range of such frame-based cameras, the performance of the aforementioned techniques may be affected by motion blur or temporal aliasing.

\begin{figure}[!t]
\begin{center}
  \includegraphics[width=\linewidth]{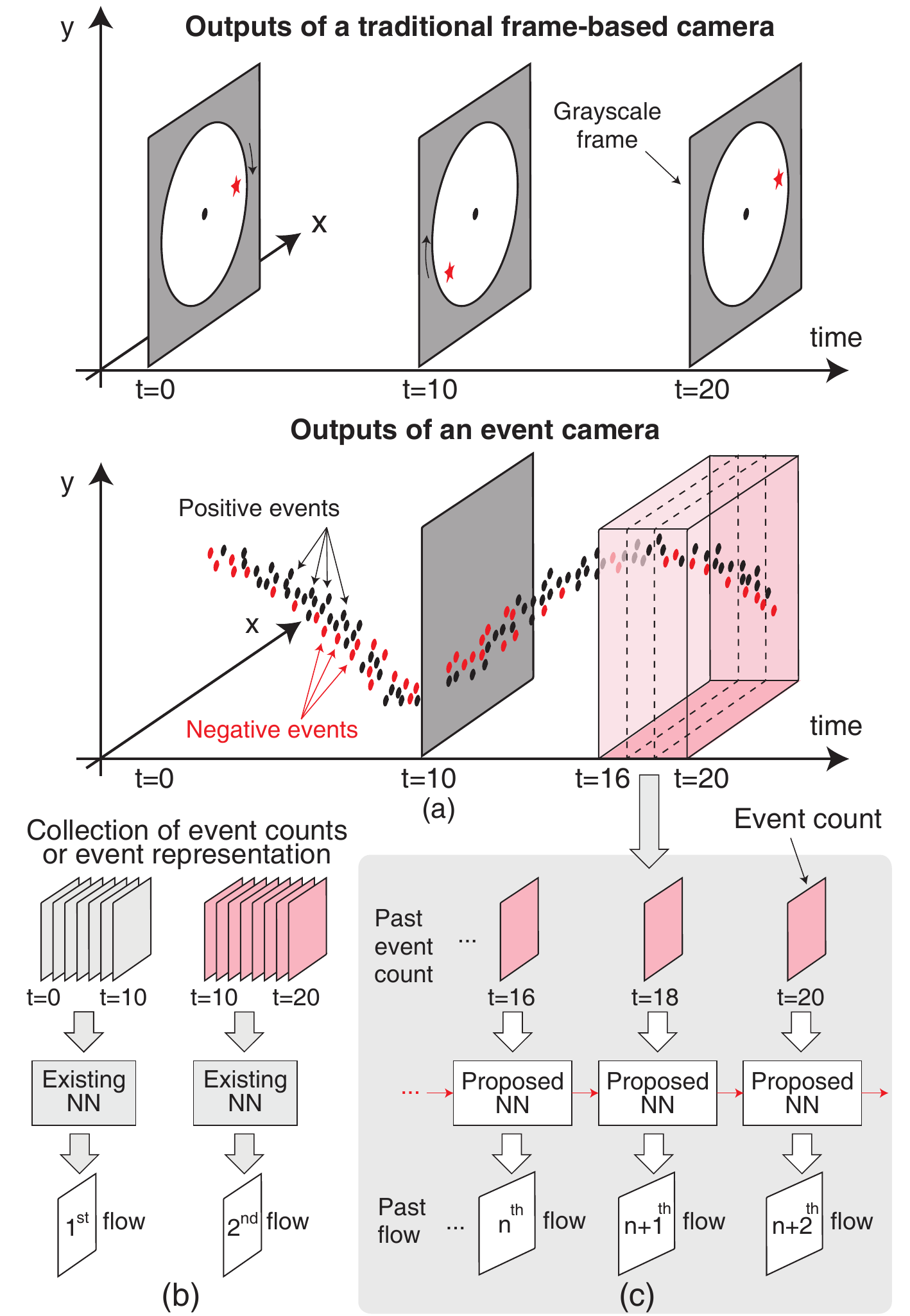}
\end{center}
\caption{(a) Comparison between outputs of a traditional frame-based and event camera. (b) Existing NN models typically rely on a collection of events for optical flow prediction. (c) We train NN models with memory elements to process each event count so that they can perform more frequent optical flow estimation. Red arrows indicate information flow from a past to a future time-step.}
\label{fig:intro_comparision}
\end{figure}

% Introduction to event-based approaches -> Drawback of the existing event-based approaches
Methods to estimate optical flow from event camera outputs offer a promising alternative to the frame-based  approaches~\cite{zhu2018ev,ye2020unsupervised,zhu2019unsupervised,lee2020spike,lee2022fusion,hagenaars2021self}.
An event camera logs light intensity change at each pixel (so-called events) rather than measuring actual light intensity for a fixed duration.  Thus, an event camera can generate a stream of events at high temporal resolution as illustrated in Fig.~\ref{fig:intro_comparision}(a). The resolution may be as small as 300~$\mu $s~\cite{gallego2020event}, making event-based optical flow estimation less susceptible to motion blur and more suitable for a highly dynamic scene.
Nonetheless, being able to effectively extract information from a high-frequency event stream is a challenging task. An event camera outputs events at a fast rate but in an asynchronous and noisy manner.  
To ensure high fidelity of the inputs to the NN models, existing works collect events over a fixed period (often a duration between two consecutive optical flow ground truths) and construct a spatio-temporal representation for optical flow estimation. 
Hence, optical flow is evaluated at a speed slower than the rate that events are produced by an event camera as illustrated in Fig.~\ref{fig:intro_comparision}(b). 
Evaluating optical flow at a faster rate can be crucial for certain applications, such as dodging an obstacle during navigation~\cite{sanket2020evdodgenet}, where fast reaction time is essential.

% Talk about the proposed approaches -> Potential benefits -> Main contributions
To predict temporally dense optical flow, we cast the event-based optical flow estimation as a sequential learning problem. We consider the event stream as a long correlated sequence over time rather than multiple independent sequences of inputs like in the existing works\cite{zhu2018ev,ye2020unsupervised,zhu2019unsupervised,lee2020spike,lee2022fusion,gehrig2021raft}.
This approach allows us to reduce the time needed to collect events as depicted in 
 Fig.~\ref{fig:intro_comparision}(c).
We train the NN models to learn the trajectory from each event count and use the collected information to estimate optical flows.
NN models are hence, required to have internal states that are capable of retaining history.
For demonstrating the efficiency of our training method, we first construct an NN model similar to the commonly used model in event-based optical flow estimation, EV-FlowNet\cite{zhu2018ev}, but replace each convolutional layer with a layer of convolutional long-short term memory (LSTM)~\cite{shi2015convolutional}. 
The use of LSTM allows previous event information to be stored and evolved through time.
To demonstrate the possibility of implementing temporally dense optical flow estimation for real-time application, we construct another NN model similar to EV-FlowNet but replace stateless neurons (like ReLU) with stateful spiking neurons~\cite{gerstner2014neuronal}.
Spiking neural networks (SNNs) have been previously proposed to address the inefficiency of typical neural networks in handling events which are sparse in nature~\cite{lee2020spike,lee2022fusion}.
Note that neurons communicate with other neurons through binary values, and hence, SNNs offer power savings on event-driven hardware by processing only non-zero inputs.
In addition, SNNs have internal states (membrane potentials) which enable them to retain information over time. This inherent recurrence in SNNs can be advantageous for sequential learning tasks such as temporally dense optical flow estimation.
We demonstrate that our training methodology can be applied to the spiking models, resulting in a model with significantly fewer parameters than the corresponding LSTM model.
% The spiking model provides substantial power savings while achieving dense temporal flow. 
Our estimation reveals that the spiking model consumes only 58\% energy compared to the baseline EV-FlowNet while predicting 10$\times$ more frequent optical flow. Successful training of the spiking model serves as the first step to realize temporally dense flow estimation on a neuromorphic chip like Intel Loihi~\cite{davies2018loihi} which recently achieved a throughput of 1000+ fps for multi-layer convolutional SNN computation~\cite{viale2021carsnn}.

Throughout this work, we refer to the two proposed models as LSTM-FlowNet and EfficientSpike-FlowNet, respectively, for short.
Steps to train both models for temporally dense optical flow estimation are, nonetheless, not straightforward.
A proper encoding scheme must be adapted to deliver event information during every small duration to the models. For this purpose, we use per-pixel event count obtained through simple aggregation over a small time period. Temporal information of the events is implicitly encoded in the order that the event counts are fed to the models.
Despite its simplicity, we show that the event count is sufficient for optical flow estimation and in fact leads to better prediction with a sequential learning methodology.
Another challenge comes from a typical assumption in sequential learning that an input has a limited length. However, an input in our case (i.e., event stream) is a long indefinite-length sequence of information, as the optical flow estimation may be performed for an extended time.
This raises the issue of how to estimate optical flow from an event stream without resetting the NN models.
Resetting the models would result in losing valuable event information processed in the past.
We find that typical sequential learning approaches do not train the models to perform well on continuous inference (i.e., without a regular reset) and propose modifications to address this problem. Our proposed modification allows the models to learn and ignore information from older events while considering more recent relevant events for optical flow estimation. 

Overall, our contributions can be summarized as follows: 
\begin{enumerate}
\item{We cast event-based optical flow estimation as a sequential learning problem to achieve temporally dense optical flow prediction. We introduce two NN models with internal states, namely LSTM-FlowNet and EfficientSpike-FlowNet, and train them on the DSEC dataset~\cite{gehrig2021dsec} to estimate 10$\times$ more frequent optical flow than models crafted from the existing approaches.}
\item{We present a technique to train the proposed models for optical flow estimation without any network reset, so that information from past relevant events is carried over time for a more reliable and frequent prediction. We show that an ability to draw longer temporal correlations from an event stream leads to 13\% improvement in the flow prediction accuracy of LSTM-FlowNet over the baseline EVFlowNet.}
\item{We demonstrate the potential of efficiently estimating more frequent flow (temporally dense flow) by applying the proposed method to train EfficientSpike-FlowNet. Compared to LSTM-FlowNet, we found that the spiking model has a higher prediction error due to its simpler recurrence dynamic. However, it comes with 3.23$\times$ lower number of parameters and offers substantial power savings (1.5\% of the LSTM-FlowNet). 
}
\end{enumerate}

\section{Background}     \label{sec:bg}
    
\subsection{Comparison with Existing Works}

The primary focus of many existing works on event-based optical flow estimation is on proposing different NN models for predicting optical flow~\cite{zhu2018ev,zhu2019unsupervised,ye2020unsupervised,lee2020spike,lee2022fusion,hagenaars2021self,gehrig2021raft,kosta2023adaptive}.
Zhu et al. proposed the first encoder-decoder model known as EV-FlowNet to process an event representation~\cite{zhu2018ev}.
A similar model was introduced in~\cite{zhu2019unsupervised,ye2020unsupervised} but with an ability to compute camera and depth simultaneously.
The inefficiency of EV-FlowNet in handling events was addressed in~\cite{lee2020spike,lee2022fusion} by incorporating spiking neurons into the encoder part of the model.
With the advancement of SNN training techniques, the fully spiking variances of EV-FlowNet were introduced later in~\cite{hagenaars2021self,kosta2023adaptive}. 
Another line of works proposed the use of recurrent NNs for iteratively optimizing optical flow~\cite{ding2022spatio,gehrig2021raft}.
A key observation is that all models proposed so far are still trained similar to the EV-FlowNet. They are trained to predict a single optical flow during a fixed interval, such as 10~Hz used in training on the DSEC dataset.
Our work distinguishes itself from prior studies by proposing a new methodology for predicting multiple optical flows within the same period.
We achieve this temporally dense optical flow estimation by leveraging proper sequential training and recurrent NNs with internal states (outlined in Section~\ref{sec:method}).
One may argue that event-based flow estimation with recurrent NNs has been proposed in~\cite{ding2022spatio,gehrig2021raft}. However, those models are designed to process each event representation multiple times and utilize that information to iteratively improve flow prediction. Our proposed models, on the other hand, are trained to process each event count only once and retain relevant information for an optical flow estimation.

\begin{figure}[!t]
\begin{center}
  \includegraphics[width=\linewidth]{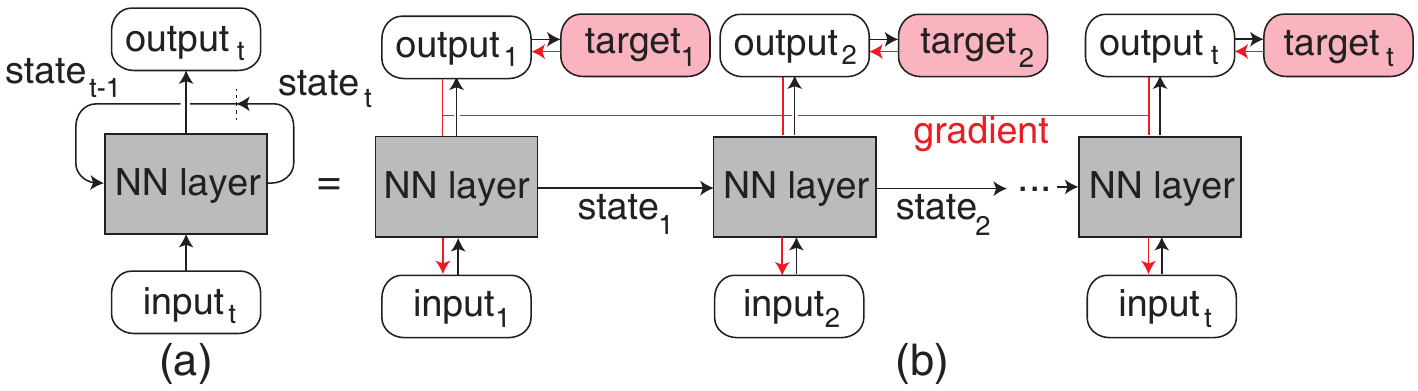}
\end{center}
\caption{(a) Diagram representing operations of LSTM or spiking neuron layer at time $t$. Internal state of both NN layers is carried over from the past time-step to the current time-step for computation. (b) Equivalent representation when operations of LSTM or spiking neuron layer are unrolled into multiple time-steps.}
\label{fig:bg_typical_seq_learning}
\end{figure}

\subsection{Building Blocks for Sequential Learning}

Sequential learning tasks are a class of problems where information is received through multiple episodes over time. In sequential learning, NNs are trained to extract and retain important information at each time step for future predictions (e.g., optical flow). 
This calls for NNs with memory elements to retain information from the past.
In this work, we utilize two different types of NN layer to create a model like EV-FlowNet, namely convolution LSTM and spiking neuron layer. The operation of convolution LSTM and spiking neuron layer can be visualized in the form of a computational graph as shown in Fig.~\ref{fig:bg_typical_seq_learning}(a).

\begin{figure}[!t]
\begin{center}
  \includegraphics[width=\linewidth]{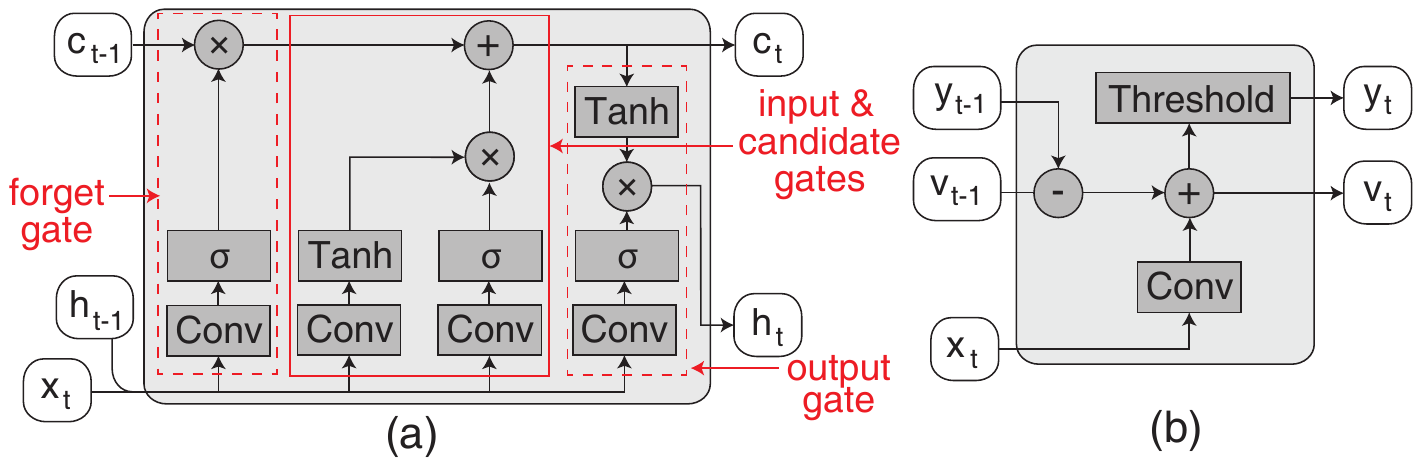}
\end{center}
\caption{Computational graph of (a) convolutional LSTM layer and (b) convolutional spiking neuron layer.}
\label{fig:bg_lstm_and_snn}
\end{figure}

\subsubsection{Convolutional LSTM}

LSTM is a type of NNs with internalized memory that have demonstrated exceptional generalization capabilities across various sequential learning problems~\cite{hochreiter1997long,graves2013speech,sutskever2014sequence}. 
Its internal state typically referred to as the cell state ($c_t$) is designed specifically to avoid the vanishing gradient problem. 
The cell state runs straight through an LSTM with minimal linear interactions as illustrated in Fig.~\ref{fig:bg_lstm_and_snn}(a), thus avoiding encoded information from the past being disrupted while sustaining gradient flow during back-propagation. LSTM can be combined with a convolutional layer making it capable of processing 2-D inputs.

Whenever a layer of convoluation LSTM receives an input ($x_t$) at time $t$, it computes an output known as the hidden state ($h_t$) and the internal state ($c_t$) through various gating mechanisms as follows. 
First, the forget gate (see left dashed box in Fig.~\ref{fig:bg_lstm_and_snn}(a)) controls how much of the previous cell state ($c_{t-1}$) is retained by deriving a scale factor $f_t$ (valued between 0 and 1) based on the input and the previous hidden state ($h_{t-1}$).
Candidate and input gates (see middle box in Fig.~\ref{fig:bg_lstm_and_snn}(a)) then calculate the contribution from the input to the internal state and combine it with the output of the forget gate to obtain the new cell state.
Lastly, the output gate controls the amount of information carried from the new cell state to the convolutional LSTM output (hidden state).
The dynamic of the convolutional LSTM layer can be expressed mathematically as follows:
\begin{align*}
    f_t &= \sigma (\text{Conv}(h_{t-1}) + \text{Conv}(x_t) + b_f) \\
    i_t &= \sigma (\text{Conv}(h_{t-1}) + \text{Conv}(x_t) + b_i) \\
    \hat{c}_t &= \text{tanh}(\text{Conv}(h_{t-1}) + \text{Conv}(x_t) + b_c) \\
    c_t &= f_t \odot c_{t-1} + i_t 	\odot \hat{c}_t \\
    o_t &= \sigma (\text{Conv}(h_{t-1}) + \text{Conv}(x_t) + b_f) \\
    h_t &= o_t \odot \text{tanh}(c_t)
\end{align*}
where $b$ are a bias of each different gate. $\odot$ signifies an element-wise multiplication.

\subsubsection{Spiking Neurons}
Spiking neurons are artificial neurons that are inspired by biological neurons in nature. They display several unique characteristics that make them suitable for real-time applications. 
Artificial spiking neurons communicate sparsely through binary signals (so-called spikes) that resemble electric pulses transmitted by biological neurons. This communication scheme simplifies hardware implementations of SNNs and enables their computations to be done efficiently in an event-driven manner~\cite{davies2018loihi,debole2019truenorth,orchard2021efficient}. 
In addition, their event-driven nature makes them ideal for handling asynchronous data generated by event sensors.
Spiking neurons also have internal states which are useful for sequential learning.

Dynamics of the leaky integrate-and-fire (LIF) neuron, a popular spiking neuron model, have a couple of notable characteristics~\cite{gerstner2014neuronal}.
A spiking neuron has an internal state referred to as the membrane potential ($v_t$).
The membrane potential is increased by an input coming into the neuron after the input gets modulated by a synaptic weight. The neuron then generates an output (or a spike) when the membrane potential exceeds a defined threshold as shown in Fig.~\ref{fig:bg_lstm_and_snn}(b).
Mathematically, the dynamics of the LIF neuron with a convolutional connection that we used in this work can be expressed as follows:
\begin{align*}
    v_t &= v_{t-1} - y_{t-1} + \text{Conv}( x_t) + b  \\
    y_t &= \text{thres}(v_t)
\end{align*}
where $x_t$ and $y_t$ represent the input and output of the LIF neuron at time $t$. $b$ is a bias of the neuron.

\subsection{Method to Train Sequential Networks}

To understand the training methodology, we refer to the computational graphs of the NN layer in Fig.~\ref{fig:bg_typical_seq_learning}. 
Because internal states of the NN layer ($state_{t}$) are computed based on new inputs ($input_t$) and their state values from the previous time-step ($state_{t-1}$), we can utilize the same computational graph to derive new internal states and outputs ($output_t$) recursively. Hence, the back-propagation through time (BPTT) algorithm can be applied to compute gradients for training these models.
For this, the operations of the NN layer are unfolded in time by creating several copies of it and treating them as a feed-forward network with tied weights.
Fig.~\ref{fig:bg_typical_seq_learning}(b) shows the computation graph after an unrolling. Given a target flow at each time, an error can be computed in a supervised manner and the gradient can be then propagated backward to each time-step.
We overcome the non-differentiability of the SNN threshold function by using surrogate gradients~\cite{zenke2018superspike,bellec2018long}.

%------------------------------------------------------------------------------
% Section 3: Method
%------------------------------------------------------------------------------

\section{Proposed Method}    \label{sec:method}

\subsection{Proposed Event Representation for Temporally Dense Flow Estimation}

Selecting a proper event representation for optical flow estimation is a challenging task as an event camera asynchronously reports changes in the light intensity ($I_t$) at every pixel on the sensor array. For each pixel, an event camera can generate a negative event or a positive event. The positive event is generated whenever the brightness increases beyond a predefined threshold ($\theta^{+}$) as described by the following equation:
\begin{align*}
    \log(I_t/I_{t-1}) \ge \theta^{+}
\end{align*}
Likewise, the negative event is generated when the brightness decreases beyond a different threshold $\theta^{-}$. Hence, each event corresponds to a time $(t)$, pixel location $(x, y)$ and polarity of change $(p)$.
Since the goal of optical flow estimation is to produce image-like output that indicates the flow magnitude in $x$-and $y$-directions, existing approaches utilize a convolutional layer to draw spatial correlations between nearby pixels. A common practice is to structure event information as frames with a fixed number of channels before convolution operation is applied. 

Prior works proposed different methods to construct this spatio-temporal representation from a collection of events. One common approach encodes the average timing or the most recent timing of events at every pixel into one of the channels to capture temporal information~\cite{zhu2018ev,ye2020unsupervised}.
Another common approach divides events into multiple partitions with the same number of events. Then, per-pixel event count from each partition is calculated to form a multi-channel input~\cite{lee2020spike,lee2022fusion,hagenaars2021self}.
The issue with such input encoding schemes is that an inference can only be made once the entire sequence of event data is available. For instance, suppose that we want to represent events received over a duration between $t{=}16$ and $t{=}20$ from an event stream as depicted in Fig.~\ref{fig:intro_comparision}(c). 
At $t{=}18$, events cannot be divided into equivolume partitions and translated to a spatio-temporal representation since the total number of events that arrive during the whole duration is not yet known.

To enable instantaneous computing from events in a smaller interval, we feed per-pixel event count as an input to NNs. This representation can be obtained through simple aggregation over each time period. Since our proposed NNs process input sequentially, temporal information of events is implicitly encoded in the order that the event counts are fed to the NNs. 
We sample event count at regular intervals to keep the notion of time consistent and allow NNs to learn temporal correlations between events at each pixel.

\subsection{Proposed Models for Temporally Dense Optical Flow Estimation}

Encoder-decoder network architecture has been widely adopted by prior works for event-based optical flow estimation~\cite{zhu2019unsupervised,lee2020spike,lee2022fusion,hagenaars2021self}. This architecture has multiple downsampling convolutional layers followed by upsampling convolutional layers. The former downsampling part of the network aims to encode spatio-temporal inputs into intermediate representations while the latter upsampling part utilizes these representations to estimate optical flow.
We follow the same convention and construct two NN models for temporally dense optical flow prediction.
To demonstrate the efficiency of our training method, we first create an NN model called LSTM-FlowNet, which is similar to EV-FlowNet - a popular encoder-decoder model for event-based optical flow estimation~\cite{zhu2018ev}. However, instead of using regular convolutional layers, we replace each layer with a layer of convolutional LSTM. The use of LSTM allows previous event information to be stored and evolved through time.
In addition, we construct another NN model similar to EV-FlowNet but with one major difference. Rather than using stateless neurons like ReLU, we replace them with stateful spiking neurons.  Our aim is to demonstrate the potential implementation of temporally dense flow estimation for real-time application.
SNNs have previously been proposed to address the inefficiency of typical neural networks in handling events that is sparse in nature. By communicating through binary values, SNNs can skip computation with zero inputs when realized on event-driven hardware, resulting in power savings. Thus, we refer to the spiking model as EfficientSpike-FlowNet and analyze its expected computation requirements in the following section to demonstrate its computational efficiency.

\begin{figure}[!t]
\begin{center}
  \includegraphics[width=\linewidth]{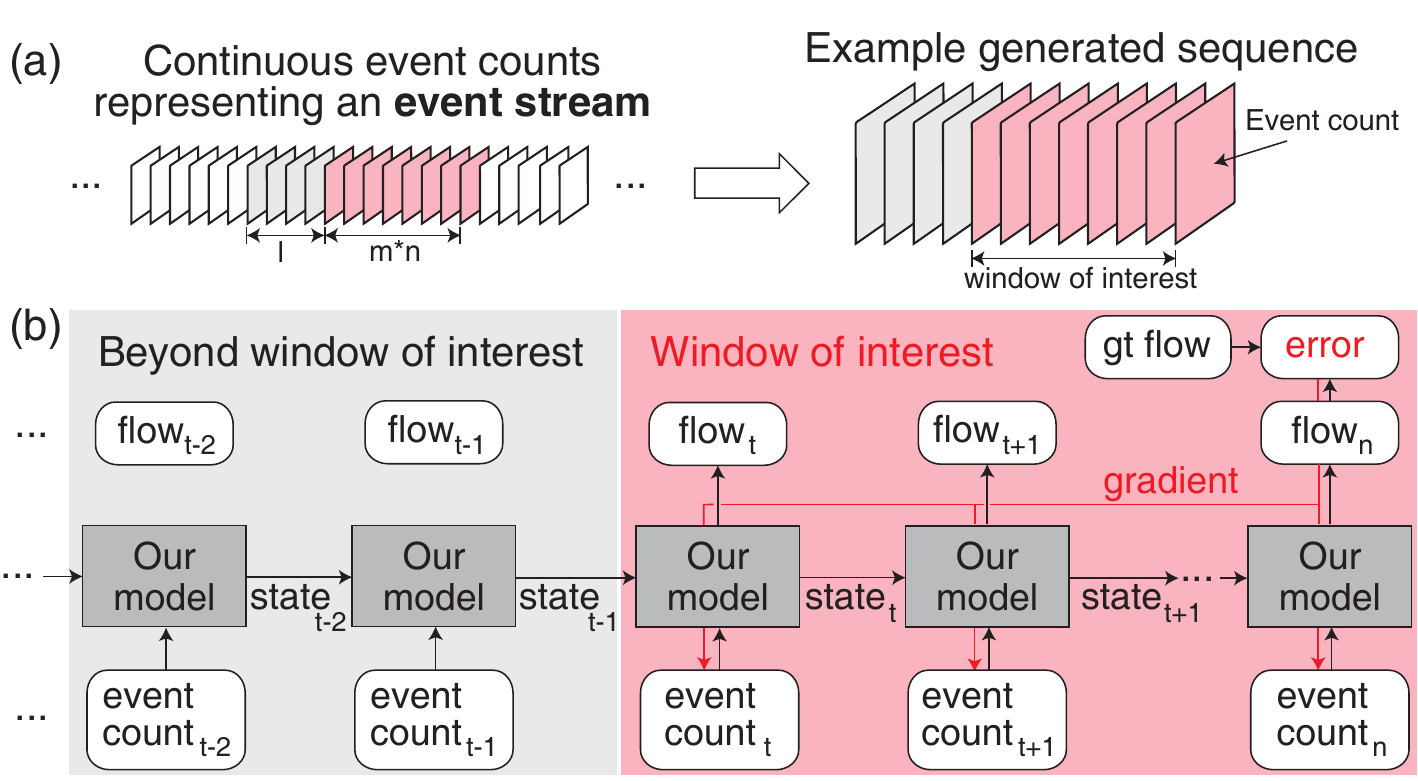}
\end{center}
\caption{(a) Example of event count sequence generation from an event stream. (b) Operations of LSTM-FlowNet and EfficientSpike-FlowNet on each generated sequence. BPTT algorithm is applied as usual, but the gradient is propagated backward only to event counts that are within a window of interest.}
\label{fig:method_our_seq_learning}
\end{figure}

\subsection{Sequential Training for Temporally Dense Optical Flow Estimation from an Event Stream} \label{sec:method_seq_training}

In order to achieve frequent optical flow estimation, we treat the event stream as one long input, rather than dividing it into individual sequences like in the previous works. However, training the proposed models on such a long input poses several challenges.
Firstly, a batch computation technique cannot be used, which leads to slower training and potential biases in the trained models. Additionally, there are limited data augmentations that can be applied during each epoch since they must be uniform across the entire input sequence.
Moreover, traditional sequential training methodology assumes that sequential inputs have a finite length, and the model's internal states are reset with each new input sequence. 
However, we want our models to estimate optical flow without interruption as reinitializing their internal states would result in a loss of information from past events.
The models also do not generate reliable outputs until they process a sufficient number of event counts. 

To address those issues, we can naively split a long event stream into multiple smaller sequences consisting of 10 event counts as illustrated in Fig.~\ref{fig:intro_comparision}(b) and train the proposed model using a typical sequential training methodology.
Then, we utilize the trained model without a network reset.
However, our preliminary experiments show that this approach results in unacceptably large errors with both proposed models, even with the use of data augmentations and noisy initial states during training.
We observe that the prediction error increases drastically after the first acceptable optical flow estimation. This is because traditional sequential training methodologies do not train the model to effectively ignore older events and focus on more recent ones. The internal states collect residual information, making the model progressively hard to estimate reliable flow with each new input.

To prepare the models for inference on an event stream without a network reset, we propose a two-step data generation approach for training.
Suppose that optical flow ground truth is available at every $m$ event counts. The first step is to create input sequences from an event stream consisting of $m{\cdot}n$ consecutive event counts (see Fig.~\ref{fig:method_our_seq_learning}(a)). We make sure that $m{\cdot}n$ is sufficiently large so that all important event counts are included for optical flow estimation. Doing so allows us to apply different data augmentations to each sequence and increases the number of data points for training. To make the model aware of the previous event counts, we increase the length of each sequence by including $l$ additional event counts in front of the $m{\cdot}n$ event counts.
The next step is to train the model using the BPTT algorithm, but propagate gradient backward only to event counts that are within a window of interest equal to $m{\cdot}n$ (see Fig.~\ref{fig:method_our_seq_learning}(b)). Information from $l$ event counts beyond this window of interest is automatically treated as noise during training.
Thus, we guarantee the models to learn temporal correlations from $m{\cdot}n$ event counts by propagating gradient back in time.

%------------------------------------------------------------------------------
% Section 4: Experimental Setting and Results
%------------------------------------------------------------------------------

\begin{figure*}[!t]
\begin{center}
  \includegraphics[width=\linewidth]{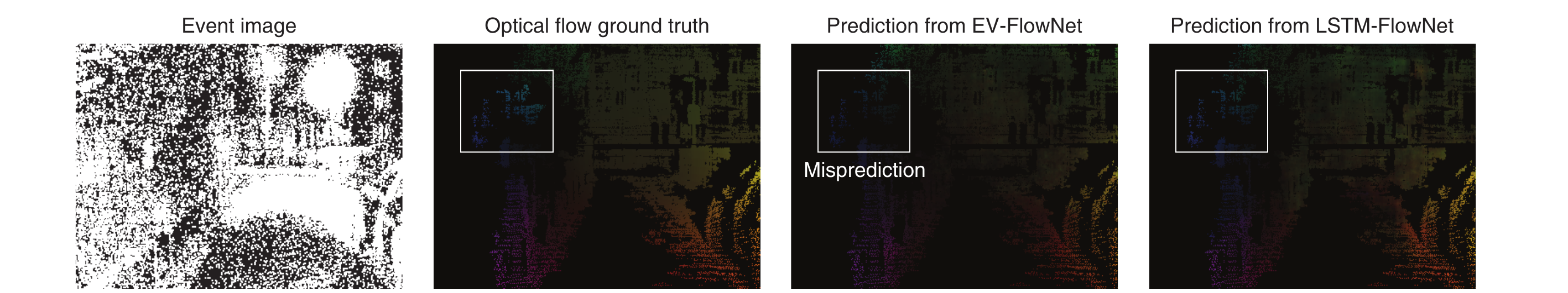}
\end{center}
\caption{Qualitative comparison of the optical flow estimated by EV-FlowNet and LSTM-FlowNet on zurich\_city\_03\_a sequence.}
\label{fig:results_visualization}
\end{figure*}

\section{Experimental Setup and Results}  \label{sec:experiment}

\subsection{Dataset, Training and Evaluation Procedure}
 
We demonstrate the effectiveness of the proposed temporally dense optical flow estimation on the DSEC dataset~\cite{gehrig2021dsec} which contains both high-resolution events and optical flow ground truths from daytime and nighttime outdoor driving under various lighting conditions.
There is another popular dataset, MVSEC~\cite{zhu2018ev}; however, the magnitudes of its optical flows are mostly smaller than 3 pixels and only about 20\% of pixels in each flow ground truth are valid. We choose to experiment with DSEC dataset as it has better quality ground truths and optical flows with 5$\times$ faster movement, allowing us to quantify the improvement with our proposed approach. 
In the DSEC dataset, the events are recorded using a stereo event camera. Optical flow ground truths are derived from odometry ground truths and are publicly available for 18 scenarios.
We split events and optical flow ground truths in each recording into a training and testing set using an 80/20 ratio. In other words, we pick the first 80\% of the events and the corresponding ground truths in each recording to be a training set while we use the rest as a testing set. Note that we provide results from using a different training-testing set-splitting strategy similar to~\cite{zhu2018ev,gehrig2021raft} are included in the supplementary document. 
For training, only events from the left camera (after applying optical correction) are used for tabulating event counts. We randomly augment the events and optical flow ground truths by flipping them along vertical and horizontal directions and cropping them down to a size of 288${\times}$384.
We train the baseline and proposed networks with Adam optimizer for 10 epochs with an initial learning rate of 5${\times}$10$^{-4}$ and a batch size of 16.
Since the optical flow ground truths in DSEC dataset are available at 10~Hz, we generate additional ground truths for training our proposed models by linear interpolation to match the input frequency. Since the constant velocity assumption might not be applicable to all motion scenarios, our approach of generating additional optical flow ground truths using linear interpolation could potentially give rise to concerns. Therefore, we also present supplementary results based on actual ground truths in the supplementary document. We trained all existing and proposed models with $\mathcal{L}_{2}$ loss that minimizes the squared differences between optical flow estimation and ground truth. The loss function can be mathematically expressed as:
\begin{align*}
    \mathcal{L} = \sum_{M}\sum_{N}{\Vert{ (u,v)_\text{prediction}- (u,v)_\text{gt}}\Vert}_{2}
\end{align*}
where $M$ is the total number of ground truths in an event stream and $N$ is the number of active pixels in the ground truth. $(u,v)$ represents optical flow magnitude along $(x,y)$ directions. 

For evaluation, we center crop events from each recording and obtain event counts of size 288${\times}$384 for optical flow estimation. We sequentially feed event counts to the proposed models one by one and obtain optical flow estimation. 
Since optical flow ground truths are available at 10~Hz, we perform another augmentation to generate extra ground truths for the test set and guarantee that the proposed model generates reliable optical flows after every input. 
We report the average of end-point errors (AEE) which is the mean of the Euclidean distance between the predicted flow and the ground truth. 
We also compute the percentage of pixels that have predicted errors greater than $k$ number of pixels (denoted as $k$PE). 
Since not all ground truth pixels are valid, we limit a calculation of these metrics to areas that have odometry information available.

\begin{table*}[!t]
\vspace{3mm}
\caption{Comparison of the flow prediction rate, average end-point error (AEE), and predicted errors greater than $k$ pixels ($k$PE) between existing and proposed models trained using different learning methodologies. \textbf{Bold} value represents the best result of each metric.}
\label{tab:perf_comparision}
\begin{center}
\begin{tabular}{llccccc} 
\hline
Training method & Architecture & Prediction rate & AEE & 1PE & 2PE & 3PE \\
\hline
Using each corresponding & EV-FlowNet~\cite{zhu2018ev} & 10 Hz & 0.67 & 17\% & 3\% & 1\% \\
event representation & Spike-FlowNet~\cite{lee2020spike} & 10 Hz & 1.12 & 64\% & 28\% & 13\% \\
constructed from events & Adaptive-FlowNet~\cite{kosta2023adaptive} & 10 Hz & 1.26 & 47\% & 15\% & 6\% \\
every 10 Hz & E-RAFT~\cite{gehrig2021raft} & 10 Hz & \textbf{0.52} & \textbf{10\%} & \textbf{2\%} & \textbf{1\%} \\
\hline
Typical sequential & LSTM-FlowNet & \textbf{100 Hz} & 36.91 & 100\% & 100\% & 100\% \\
learning method & EfficientSpike-FlowNet & \textbf{100 Hz} & 20.99 & 100\% & 99\% & 99\% \\
\hline
Proposed sequential & LSTM-FlowNet & \textbf{100 Hz} & 0.60 & 12\% & \textbf{2\%} & \textbf{1\%} \\
learning method & EfficientSpike-FlowNet & \textbf{100 Hz} & 2.66 & 84\% & 56\% & 34\% \\
\hline
\end{tabular}
\end{center}
\end{table*}

\subsection{Optical Flow Estimation Rate and Accuracy}

Table~\ref{tab:perf_comparision} presents a comparison of the optical flow estimation rate, AEE, and $k$PE between the existing and proposed models.
To train the existing models, we first use their corresponding event representations proposed in each work as inputs to the models.
As the optical flow ground truths on the DSEC dataset are recorded at 10Hz, we split the event stream at times when ground truths are available and construct a spatio-temporal representation based on events in each split. 
Existing models are trained with the event representations at 10~Hz, resulting in models that estimate optical flow at the same frequency (see column 3 of the first four rows in the table). In contrast, we train LSTM-FlowNet and EfficientSpike-FlowNet using the proposed event representation, which enables more frequent optical flow estimation. The proposed models receive event counts which can be computed during a much shorter interval since normalization or other pre-processing is not required. 
We arbitrarily collect event counts at 100~Hz, which is 10 times of the optical flow ground truth frequency. Other rates are possible as discussed in the following subsection. We then utilize sequential training method to train the proposed models for temporally dense optical flow at the rate of 100 Hz. As a result, the prediction rate for the proposed models is an order of magnitude higher than the existing ones (see column 3 of the last four rows in the table). 
Note that we evaluate both proposed models without a network reset to reproduce a scenario where the models are used for real-time optical flow estimation. Network reset implies an interruption in getting reliable optical flows as a sequential model requires processing a sufficient number of event counts similar to the way it was trained before producing a faithful prediction.

\begin{table*}[!t]
\caption{Comparison of the optical flow prediction rate, AEE, number of parameters, and normalized compute energy per second between baseline and proposed models with different types of inputs. \textbf{Bold} value represents the best result of each metric.}
\label{tab:eff_comparision}
\begin{center}
\begin{tabular}{llcccc}
\hline
Inputs & Architecture & Prediction rate & AEE & \# Params & Normalized energy \\
\hline\
Event representation at 10 Hz & EV-FlowNet~\cite{zhu2018ev} & 10 Hz & 0.67 & \textbf{16.6M} & 1$\times$ \\
% \multirow{2}{*}{Event representation at 10 Hz} & EV-FlowNet~\cite{zhu2018ev} & \multirow{2}{*}{10 Hz} & 0.67 & 16.6M & 1$\times$ \\
%  & E-RAFT~\cite{gehrig2021raft} & & \textbf{0.52} & \textbf{5.3M} & 2.5$\times$ \\
\hline
\multirow{2}{*}{Event counts at 100 Hz} & LSTM-FlowNet & \multirow{2}{*}{\textbf{100 Hz}} & \textbf{0.60} & 53.6M & $40\times$ \\
& EfficientSpike-FlowNet & & 2.66 & \textbf{16.6M} & 0.58$\times$ \\
\hline
Event counts at 50 Hz & EfficientSpike-FlowNet & 50 Hz & 3.86 & \textbf{16.6M} & \textbf{0.24$\times$} \\
\hline
\end{tabular}
\end{center}
\end{table*}

Our results reveal that a typical sequential learning method does not train the proposed models well for optical flow estimation without a regular state reinitialization as discussed in Section~\ref{sec:method_seq_training}. Both the proposed models perform poorly in terms of prediction accuracy (see column 4 of the middle two rows in the table).
Our proposed sequential training method (with $n$=1 and $l$=10) addresses this potential issue in optical flow estimation. It enables LSTM-FlowNet and EfficientSpike-FlowNet to estimate temporally dense optical flows with a mean error smaller than the average flow magnitude in our testing set (7.73 pixels).
LSTM-FlowNet in particular outperforms all existing models with encoder-decoder architecture. 
Compared to the baseline EV-FlowNet, LSTM-FlowNet achieves a 13\% lower AEE, thanks to its ability to draw longer correlations back in time. Our qualitative comparison reveals that LSTM-FlowNet outperforms EV-FlowNet in scenarios with only few reliable events such as events generated from a tree line under low illumination as shown in Fig.~\ref{fig:results_visualization}. Nonetheless, LSTM-FlowNet still has slightly lower accuracy than E-RAFT which relies on a different model architecture and principle. 1PE measurement 
%% what is 1PE? Reference?
indicates that their differences come from the predicted flows with errors of 1 pixel or less~\cite{gehrig2021raft}.
% The increase in the flow prediction accuracy of both LSTM-FlowNet and E-RAFT comes with an additional computational cost as we discussed in the following subsection.
While EfficientSpike-FlowNet benefits from the proposed sequential learning method, its flow estimation accuracy is slightly lower than LSTM-FlowNet,
%% is it correct? accuracy is higher for EfficientSpikeFlowNet?
which has more complex recurrent dynamics. However, we show that the simple recurrent dynamics turn out to be beneficial in terms of the number of parameters (see column 5 of Table~\ref{tab:eff_comparision}). The spiking model has 3.23$\times$ lower number of parameters than LSTM-FlowNet, which translates to smaller memory requirements and potentially lower power consumption. 

\subsection{Computational Efficiency}

We evaluate the computational efficiency of the baseline and proposed models by measuring the expected energy consumption as shown in the last column of Table~\ref{tab:eff_comparision}. To compute the energy consumption, we adopt a similar approach to that used in~\cite{lee2022fusion,rueckauer2017conversion}, which calculates energy based on the number and type of arithmetic operations.
Since spiking neurons communicate through binary values (0 and 1), power-hungry multiplication operations in SNN can be simplified into addition operations. 
For computing weight sum, SNNs perform sparse accumulate (AC) operations instead of multiply-and-accumulate (MAC) operations used by typical NNs. The energy required for AC and MAC operations in 32-bit floating-point computation on 45nm CMOS technology are 0.9 pJ and 4.6 pJ~\cite{horowitz20141}, respectively.
%% in what technology? 45nm?
%% -> (Mint) Thanks for pointing it out. I just confirm that the technology is 45nm. This is based on the ISSCC paper that Adarsh and Chankyu refer to.
This makes arithmetic operations for SNNs roughly five times more energy-efficient than typical NNs. On event-driven hardware, SNNs also provide extra power saving by processing only non-zero inputs. To compute the total energy of EfficientSpike-FlowNet, we then track the percentage of non-zero inputs received by spiking neurons in each layer and multiply the percentage with the number of arithmetic operations to get the total energy. 
Our measurement reveals that the input sparsity (i.e., the number of zero inputs) of encoder blocks in EfficientSpike-FlowNet increases with depth and reaches a maximum of 87\% in the last encoder block. The input sparsity then gradually decreases in the decoder blocks, possibly due to a reduction in the number of decoder channels.
We found that the compute energy of EfficientSpike-FlowNet is only 58\% of the baseline EV-FlowNet even though it produces more frequent optical flows. The estimated energy for EfficientSpike-FlowNet is almost two orders of magnitude lower than LSTM-FlowNet (see row 3-4 of Table~\ref{tab:eff_comparision}) due to a smaller number of parameters and its efficiency in handling events. These findings serve as a verification and represent a step towards the realization of temporally dense flow estimation on hardware geared toward fast and efficient computing like Intel Loihi~\cite{davies2018loihi} which has recently achieved a throughput of 1000+ fps for multi-layer convolutional SNN computation~\cite{viale2021carsnn}. 

\subsection{Effect of Input Rate}

In our framework, the proposed models are trained to estimate optical flow at the same frequency as the input event counts. The frequency of event counts can be changed to accommodate computational constraints.
We demonstrate that different input rates can be used by feeding the model with event counts at 50 Hz. The slower input rate results in faster training and less energy consumption (see the last row in Table~\ref{tab:eff_comparision}), as the inputs at 50~Hz require fewer computations than ones at 100~Hz within a given period. However, the error in flow estimation increases due to imprecise temporal information (i.e., using longer time to collect event count). Increasing the input rate is also possible but at the expense of inference energy consumption. In our experiments, we found that increasing the input rate beyond 100~Hz does not significantly improve the predicted flow quality. Therefore, we choose the input rate of 100~Hz in all experiments. Nonetheless, the input rate must be chosen carefully to satisfy the reaction time and computational constraints during a deployment.

%------------------------------------------------------------------------------
% Section 5. Conclusion
%------------------------------------------------------------------------------

\section{Conclusion} \label{sec:conclusion}

In this work, we propose an approach to achieve temporally dense optical flow estimation using event cameras. We cast the problem as a sequential learning task and introduce variants of the EV-FlowNet architecture that incorporate LSTMs and spiking neurons so that the models have suitable memories for learning.
Our results suggest that traditional training methods are not well-suited for training the proposed models to estimate optical flows from a continuous event stream. To address this issue, we propose a sequential training method that enables the models to focus on recent events while ignoring irrelevant older ones. This leads to a continuous 10$\times$ temporally dense flow estimation (without requiring a network reset) over existing approaches.
Results from the LSTM model reveal a potential accuracy improvement over the baseline model from the ability to draw longer temporal correlations from event streams.
We demonstrate that the inherent recurrent dynamics of the spiking model are also useful for estimating more frequent optical flow. Due to its simpler dynamics, the spiking model offers substantial parameter reduction over the LSTM model. 
In addition, our energy estimation indicates that the spiking model is significantly more efficient in handling events compared to the LSTM model, with an expected energy consumption of only 1.5\% of the LSTM one. This highlights the potential use of the spiking model for temporally dense optical flow estimation in real-time applications like flying drones with limited energy budget.

{\small
\bibliographystyle{ieee_fullname}
\bibliography{refs}

\begin{thebibliography}{10}\itemsep=-1pt

\bibitem{beauchemin1995computation}
Steven~S. Beauchemin and John~L. Barron.
\newblock The computation of optical flow.
\newblock {\em ACM computing surveys (CSUR)}, 27(3):433--466, 1995.

\bibitem{bellec2018long}
Guillaume Emmanuel~Fernand Bellec, Darjan Salaj, Anand Subramoney, Robert
  Legenstein, and Wolfgang Maass.
\newblock Long short-term memory and learning-to-learn in networks of spiking
  neurons.
\newblock In {\em Advances in Neural Information Processing Systems: NeurIPS}.
  2018.

\bibitem{davies2018loihi}
Mike Davies, Narayan Srinivasa, Tsung-Han Lin, Gautham Chinya, Yongqiang Cao,
  Sri~Harsha Choday, Georgios Dimou, Prasad Joshi, Nabil Imam, Shweta Jain,
  et~al.
\newblock Loihi: A neuromorphic manycore processor with on-chip learning.
\newblock {\em IEEE Micro}, 38(1):82--99, 2018.

\bibitem{debole2019truenorth}
Michael~V DeBole, Brian Taba, Arnon Amir, Filipp Akopyan, Alexander
  Andreopoulos, William~P Risk, Jeff Kusnitz, Carlos~Ortega Otero, Tapan~K
  Nayak, Rathinakumar Appuswamy, et~al.
\newblock {TrueNorth}: Accelerating from zero to 64 million neurons in 10
  years.
\newblock {\em Computer}, 52(5):20--29, 2019.

\bibitem{ding2022spatio}
Ziluo Ding, Rui Zhao, Jiyuan Zhang, Tianxiao Gao, Ruiqin Xiong, Zhaofei Yu, and
  Tiejun Huang.
\newblock Spatio-temporal recurrent networks for event-based optical flow
  estimation.
\newblock In {\em Proceedings of the AAAI Conference on Artificial
  Intelligence}, volume~36, pages 525--533, 2022.

\bibitem{dosovitskiy2015flownet}
Alexey Dosovitskiy, Philipp Fischer, Eddy Ilg, Philip Hausser, Caner Hazirbas,
  Vladimir Golkov, Patrick Van Der~Smagt, Daniel Cremers, and Thomas Brox.
\newblock {FlowNet}: Learning optical flow with convolutional networks.
\newblock In {\em Proceedings of the IEEE international conference on computer
  vision}, pages 2758--2766, 2015.

\bibitem{gallego2020event}
Guillermo Gallego, Tobi Delbr{\"u}ck, Garrick Orchard, Chiara Bartolozzi, Brian
  Taba, Andrea Censi, Stefan Leutenegger, Andrew~J Davison, J{\"o}rg Conradt,
  Kostas Daniilidis, et~al.
\newblock Event-based vision: A survey.
\newblock {\em IEEE transactions on pattern analysis and machine intelligence},
  44(1):154--180, 2020.

\bibitem{gehrig2021dsec}
Mathias Gehrig, Willem Aarents, Daniel Gehrig, and Davide Scaramuzza.
\newblock {DSEC}: A stereo event camera dataset for driving scenarios.
\newblock {\em IEEE Robotics and Automation Letters}, 6(3):4947--4954, 2021.

\bibitem{gehrig2021raft}
Mathias Gehrig, Mario Millh{\"a}usler, Daniel Gehrig, and Davide Scaramuzza.
\newblock {E-RAFT}: Dense optical flow from event cameras.
\newblock In {\em 2021 International Conference on 3D Vision (3DV)}, pages
  197--206. IEEE, 2021.

\bibitem{gerstner2014neuronal}
Wulfram Gerstner, Werner~M Kistler, Richard Naud, and Liam Paninski.
\newblock {\em Neuronal dynamics: From single neurons to networks and models of
  cognition}.
\newblock Cambridge University Press, 2014.

\bibitem{graves2013speech}
Alex Graves, Abdel-rahman Mohamed, and Geoffrey Hinton.
\newblock Speech recognition with deep recurrent neural networks.
\newblock In {\em 2013 IEEE international conference on acoustics, speech and
  signal processing}, pages 6645--6649. IEEE, 2013.

\bibitem{hagenaars2021self}
Jesse Hagenaars, Federico Paredes-Vall{\'e}s, and Guido De~Croon.
\newblock Self-supervised learning of event-based optical flow with spiking
  neural networks.
\newblock {\em Advances in Neural Information Processing Systems},
  34:7167--7179, 2021.

\bibitem{hochreiter1997long}
Sepp Hochreiter and J{\"u}rgen Schmidhuber.
\newblock Long short-term memory.
\newblock {\em Neural computation}, 9(8):1735--1780, 1997.

\bibitem{horn1981determining}
Berthold~KP Horn and Brian~G Schunck.
\newblock Determining optical flow.
\newblock {\em Artificial intelligence}, 17(1-3):185--203, 1981.

\bibitem{horowitz20141}
Mark Horowitz.
\newblock 1.1 computing's energy problem (and what we can do about it).
\newblock In {\em 2014 IEEE International Solid-State Circuits Conference
  Digest of Technical Papers (ISSCC)}, pages 10--14. IEEE, 2014.

\bibitem{janai2020computer}
Joel Janai, Fatma G{\"u}ney, Aseem Behl, Andreas Geiger, et~al.
\newblock Computer vision for autonomous vehicles: Problems, datasets and state
  of the art.
\newblock {\em Foundations and Trends{\textregistered} in Computer Graphics and
  Vision}, 12(1--3):1--308, 2020.

\bibitem{kosta2023adaptive}
Adarsh~Kumar Kosta and Kaushik Roy.
\newblock {Adaptive-SpikeNet}: Event-based optical flow estimation using
  spiking neural networks with learnable neuronal dynamics.
\newblock In {\em 2023 International Conference on Robotics and Automation
  (ICRA)}. IEEE, 2023.

\bibitem{lee2022fusion}
Chankyu Lee, Adarsh~Kumar Kosta, and Kaushik Roy.
\newblock {Fusion-FlowNet}: Energy-efficient optical flow estimation using
  sensor fusion and deep fused spiking-analog network architectures.
\newblock In {\em 2022 International Conference on Robotics and Automation
  (ICRA)}, pages 6504--6510. IEEE, 2022.

\bibitem{lee2020spike}
Chankyu Lee, Adarsh~Kumar Kosta, Alex~Zihao Zhu, Kenneth Chaney, Kostas
  Daniilidis, and Kaushik Roy.
\newblock {Spike-FlowNet}: event-based optical flow estimation with
  energy-efficient hybrid neural networks.
\newblock In {\em European Conference on Computer Vision}, pages 366--382.
  Springer, 2020.

\bibitem{orchard2021efficient}
Garrick Orchard, E~Paxon Frady, Daniel Ben~Dayan Rubin, Sophia Sanborn,
  Sumit~Bam Shrestha, Friedrich~T Sommer, and Mike Davies.
\newblock Efficient neuromorphic signal processing with loihi 2.
\newblock In {\em 2021 IEEE Workshop on Signal Processing Systems (SiPS)},
  pages 254--259. IEEE, 2021.

\bibitem{quan2021holistic}
Ruijie Quan, Linchao Zhu, Yu Wu, and Yi Yang.
\newblock Holistic {LSTM} for pedestrian trajectory prediction.
\newblock {\em IEEE transactions on image processing}, 30:3229--3239, 2021.

\bibitem{ranjan2017optical}
Anurag Ranjan and Michael~J Black.
\newblock Optical flow estimation using a spatial pyramid network.
\newblock In {\em Proceedings of the IEEE conference on computer vision and
  pattern recognition}, pages 4161--4170, 2017.

\bibitem{rueckauer2017conversion}
Bodo Rueckauer, Iulia-Alexandra Lungu, Yuhuang Hu, Michael Pfeiffer, and
  Shih-Chii Liu.
\newblock Conversion of continuous-valued deep networks to efficient
  event-driven networks for image classification.
\newblock {\em Frontiers in neuroscience}, 11:682, 2017.

\bibitem{sanket2020evdodgenet}
Nitin~J Sanket, Chethan~M Parameshwara, Chahat~Deep Singh, Ashwin~V
  Kuruttukulam, Cornelia Ferm{\"u}ller, Davide Scaramuzza, and Yiannis
  Aloimonos.
\newblock {EvDodgeNet}: Deep dynamic obstacle dodging with event cameras.
\newblock In {\em 2020 IEEE International Conference on Robotics and Automation
  (ICRA)}, pages 10651--10657. IEEE, 2020.

\bibitem{serres2017optic}
Julien~R Serres and Franck Ruffier.
\newblock Optic flow-based collision-free strategies: From insects to robots.
\newblock {\em Arthropod structure \& development}, 46(5):703--717, 2017.

\bibitem{shi2015convolutional}
Xingjian Shi, Zhourong Chen, Hao Wang, Dit-Yan Yeung, Wai-Kin Wong, and
  Wang-chun Woo.
\newblock Convolutional lstm network: A machine learning approach for
  precipitation nowcasting.
\newblock {\em Advances in neural information processing systems}, 28, 2015.

\bibitem{singh1991optic}
Ajit Singh.
\newblock {\em Optic flow computation: a unified perspective}, volume~3.
\newblock IEEE computer society press Los Alamitos, 1991.

\bibitem{sun2018pwc}
Deqing Sun, Xiaodong Yang, Ming-Yu Liu, and Jan Kautz.
\newblock {PWC}-net: {CNNs} for optical flow using pyramid, warping, and cost
  volume.
\newblock In {\em Proceedings of the IEEE conference on computer vision and
  pattern recognition}, pages 8934--8943, 2018.

\bibitem{sutskever2014sequence}
Ilya Sutskever, Oriol Vinyals, and Quoc~V Le.
\newblock Sequence to sequence learning with neural networks.
\newblock {\em Advances in neural information processing systems}, 27, 2014.

\bibitem{viale2021carsnn}
Alberto Viale, Alberto Marchisio, Maurizio Martina, Guido Masera, and Muhammad
  Shafique.
\newblock Carsnn: An efficient spiking neural network for event-based
  autonomous cars on the loihi neuromorphic research processor.
\newblock In {\em 2021 International Joint Conference on Neural Networks
  (IJCNN)}, pages 1--10. IEEE, 2021.

\bibitem{ye2020unsupervised}
Chengxi Ye, Anton Mitrokhin, Cornelia Ferm{\"u}ller, James~A Yorke, and Yiannis
  Aloimonos.
\newblock Unsupervised learning of dense optical flow, depth and egomotion with
  event-based sensors.
\newblock In {\em 2020 IEEE/RSJ International Conference on Intelligent Robots
  and Systems (IROS)}, pages 5831--5838. IEEE, 2020.

\bibitem{zenke2018superspike}
Friedemann Zenke and Surya Ganguli.
\newblock Superspike: Supervised learning in multilayer spiking neural
  networks.
\newblock {\em Neural computation}, 30(6):1514--1541, 2018.

\bibitem{zhu2018ev}
Alex~Zihao Zhu and Liangzhe Yuan.
\newblock {EV-FlowNet}: Self-supervised optical flow estimation for event-based
  cameras.
\newblock In {\em Robotics: Science and Systems}, 2018.

\bibitem{zhu2019unsupervised}
Alex~Zihao Zhu, Liangzhe Yuan, Kenneth Chaney, and Kostas Daniilidis.
\newblock Unsupervised event-based learning of optical flow, depth, and
  egomotion.
\newblock In {\em Proceedings of the IEEE/CVF Conference on Computer Vision and
  Pattern Recognition}, pages 989--997, 2019.

\end{thebibliography}
}

\end{document}